\documentclass{singlecol-new}
\usepackage{amssymb}
\usepackage{amsmath} 
\usepackage{graphicx}
\usepackage{subfigure}
\usepackage{natbib}
\usepackage{multirow}
\usepackage{url}
\usepackage{natbib}

\theoremstyle{TH}{

}

\theoremstyle{THhit}{

}

\theoremstyle{THrm}{

}

\makeatletter
\def\theequation{\arabic{equation}}
\makeatother

\begin{document}

\setcounter{page}{1}

\LRH{M. C Hanumantharaju et al.}

\RRH{PSO Based Color Image Enhancement}

\VOL{x}

\ISSUE{x}

\BottomCatch

\PAGES{xxxx}

\CLline

\PUBYEAR{2009}

\subtitle{}

\title{A New Framework for Retinex based Color Image Enhancement using Particle Swarm Optimization}

\authorA{M. C Hanumantharaju*\vs{-3}}
\affA{Department of Information Science \&\ Engineering,\\ Dayananda Sagar College of Engineering,\\
Bangalore - 560078 \\ E-mail:
mchanumantharaju@gmail.com\\ {*}Corresponding author\vs{-2}}

\authorB{M. Ravishankar\vs{-7}}
\affB{Department of Information Science \&\ Engineering,\\ Dayananda Sagar College of Engineering,\\
Bangalore - 560078 \\
E-mail: ravishankarmcn@gmail.com}

\authorC{D. R Rameshbabu\vs{-7}}
\affC{Department of Computer Science \&\ Engineering,\\ Dayananda Sagar College of Engineering,\\
Bangalore - 560078 \\
E-mail: bobrammysore@gmail.com}

\authorD{V. N Manjunath Aradhya\vs{-7}}
\affD{Department of Information Science \&\ Engineering,\\ Dayananda Sagar College of Engineering,\\
Bangalore - 560078 \\
E-mail: aradhya.mysore@gmail.com}

\begin{abstract}

A new approach for tuning the parameters of MultiScale Retinex (MSR) based color image enhancement algorithm using a popular optimization method, namely, Particle Swarm Optimization (PSO) is presented in this paper. The image enhancement using MSR scheme heavily depends on parameters such as Gaussian surround space constant, number of scales, gain and offset etc. Selection of these parameters, empirically and its application to MSR scheme to produce inevitable results are the major blemishes. The method presented here results in huge savings of computation time as well as improvement in the visual quality of an image, since the PSO exploited maximizes the MSR parameters. The objective of PSO is to validate the visual quality of the enhanced image iteratively using an effective objective criterion based on entropy and edge information of an image. The PSO method of parameter optimization of MSR scheme achieves a very good quality of reconstructed images, far better than that possible with the other existing methods. Finally, the quality of the enhanced color images obtained by the proposed method are evaluated using novel metric, namely, Wavelet Energy (WE). The experimental results presented show that color images enhanced using the proposed scheme are clearer, more vivid and efficient.

\end{abstract}

\KEYWORD{Particle Swarm Optimization; Parameter Tuning; Multiscale Retinex; Color Image Enhancement; Wavelet Energy}

\maketitle

\section{Introduction}

The pre-processing or pixel level processing is the key step in digital image processing [\cite{Jain89}] for further analysis and processing of digital images. The images obtained from capturing device exhibit several defects such as non-uniform illumination, sampling noise, poor contrast, and many others. The pre-processing of such defected images plays an important role in an image processing system. The major pre-processing operations [\citet{Gonzalez08}] generally used are as follows: image enhancement for improving the visual quality of an image, thresholding to reduce gray scale image to a binary image, filtering for noise reduction and segmentation to separate various components of an image. Among these operations, image enhancement plays an important role in the pre-processing phase of the image processing. Nowadays color image enhancement [\citet{Zhang10}] has become significant research area due to the widespread use of color images in many applications. 

The goal of an image enhancement algorithm is to acquire the finer details of an image and highlight the useful information that are not clearly visible owing to different illumination conditions. Conventional methods employed for image enhancement are Histogram Equalization (HE) [\citet{Sundaram11}], Adaptive HE [\citet{Elisabeta11}], gamma correction [\citet{Guan09}] and homomorphic filtering [\citet{Ming06}]. Although good enhancement results have been obtained from these image enhancement methods, visual inspection of the reconstructed image reveals that these enhancement schemes heavily depend on input images. On the other hand, conventional enhancement methods need manual adjustment of parameters in order to obtain acceptable results. Numerous retinex based image enhancement algorithms were developed by [\citet{Rahman}] to enhance medical, natural, facial and aerial images etc. The software based image enhancement programs such as Adobe Photoshop, Photoflair, Amped and ImaQuest etc., offers a number of interactive tools to improve the quality of the image. However, the standard imposed on the enhancement technique differs from one area of application to another.

PSO is one of the searching algorithm that can be applied to non-linear and discrete optimization problems. Numerous researchers have incorporated the PSO algorithm to solve various problems of image processing [\citet{Braik07}]. In Ref. [\citet{Mohamed10}] authors used PSO to optimize the scaling factor of linear unsharp masking based edge enhancement. [\citet{Sengee10}] used PSO to improve the clarity of ridges in fingerprint images. [\citet{SUN08}] used Quantum-behaved PSO (QPSO) scheme to improve the adaptability and versatility of image enhancement. Many authors have used PSO technique for tuning the parameters of image enhancement methods. The answer to the question when should we use swarm to solve the problems was described by [\citet{Blackwell06}]. Therefore, it is clear that tuning the parameters using PSO to achieve the required goals in image enhancement is a challenging problem. In this paper, PSO has been explited to adjust the parameters of retinex based color image enhancement. The efficiency and superiority of the proposed color image enhancement method using PSO algorithm can be confirmed by the WE metric.

The paper is organized as follows: Section 2 gives a brief review of related work. Section 3 describes the proposed particle swarm optimization based MSR with modified color restoration algorithm. Section 4 provides experimental results and comparative study. Conclusion arrived at are presented in the section 5. 

\section{Related Work}


In 1971, Edwin H. Land formulated the retinex theory. According to Land's theory, the word "retinex" is a portmanteau formed from "retina" and "cortex" suggesting that both the eye and the brain are involved in the processing. [\citet{Brainard86}] proposed a theoretical study of retinex algorithm based on convergence properties of Land's retinex theory and exhibited that the pixel values converge to simple normalization when both path and length keep increasing. However, the retinex path based methods have high computational complexity. [\citet{Funt04}] proposed an iterative version of retinex algorithm which uses 2D extension of path version. In this work, a new value for each pixel is computed by iteratively comparing pixels in an image. However, the number of iterations is not defined and thus affects the final results.  [\citet{Hyun08}] proposed the Single-Scale Retinex (SSR) model in order to enhance the color image. The SSR algorithm developed in this method relies on the ratio of the lightness of a small central field in the region of interest to the average lightness of an extended field. The Gaussian function has been used in this scheme in order to obtain the average lightness of the entire image. However, the SSR algorithm is unable to remove halo artifacts in the enhanced image completely. 
	 
[\citet{Jobson97}], proposed MSR based color image enhancement algorithm (popularly called as NASA's retinex method) in order to overcome the drawback of SSR. In this method, an input image is processed several times using SSR with the Gaussian filters of various scales. The resulting images are weighted and then summed to reduce the halo artifacts and improve the local contrast. However, this scheme may not optimize the parameters such as size and weight of the Gaussian filters. As a result, ratios for the variations in luminance and chrominance are unmatched. This mismatch inevitably leads to an unnatural color rendition. [\citet{Hongqing10}] proposed an improved retinex image enhancement algorithm. The algorithm has very good performance in color constancy, contrast and computational cost. However, the retinex based image enhancement is achieved in HSV color space in place of RGB color space. Therefore, an additional color space conversion increases the computational complexity. [\citet{Terai09}] proposed a retinex model for color image contrast enhancement. The luminance signal is processed to reduce the computation time without changing color components. Although this scheme offers satisfactory results for gray images, computational complexity of the algorithm increases due to large scale Gaussian filtering. 

[\citet{Jharna11}] proposed a retinex algorithm for color image enhancement with reduced halo artifacts. The method was inspired based on the work proposed by [\citet{Moore91}]. [\citet{Ruibo11}] proposed an algorithm for ultrasound liver image enhancement based on MSR theory. The images processed using MSR algorithm, enhances the contrast ratio and increases entropy information. This scheme offers enriched visual quality detail by improving the details in dark areas. The improved visual quality assists doctors for classification and identification of liver cells. However, the issues such as selection of environmental function and its standard deviation are to be considered in order to obtain satisfactory results. [\citet{Liu10}] proposed multi-dimensional multiscale image enhancement algorithm based on MSR theory combined with conduction function. This method provides good color constancy and better contrast enhancement over other enhancement approaches. However, an image with less affected by the illumination require larger weights and scales. Further, if the image is encountered with a dim light source requires a change of Gaussian scales and weights depending upon the source of illumination.


[\citet{Apurba09}] proposed gray level image enhancement using PSO. In this method, image enhancement is considered as an optimization problem. The solution to this problem has been obtained by exploiting PSO. Image enhancement is achieved by maximizing the information content in the enhanced image using intensity transformation. The parameterized transformation function adapted in this work uses local and global information of the image. The objective criterion of this algorithm is based on the measurement of entropy and edge information. The parameters used in the transformation function are optimized based on PSO in accordance with the objective criterion. This scheme produces better result compared with other image enhancement techniques such as contrast stretching, HE and Genetic Algorithm (GA) based image enhancement.  

[\citet{Tiedong08}] proposed sonar image enhancement based on PSO. Sonar images have important characteristics such as low resolution, strong echo disturbance, small object regions and obscure object edges. In such situations, it is very hard to obtain satisfactory results using global image enhancement algorithms. An adaptive local enhancement algorithm proposed in this work, evaluates the enhanced image using number of edges, edge intensities and the entropy value. The PSO technique has been adopted in order to search the optimal parameters for the best enhancement. Image enhancement technique based on improving the PSO algorithm has been proposed by Qunqing et al. [\citet{Qinqing11}]. In this work, image enhancement is considered as an optimization problem which optimizes the parameter using the Simulated Annealing PSO (SAPSO) method. Although, the enhancement achieved is superior than the Genetic Algorithm (GA) based image enhancement, searching curve for SAPSO is not smooth, there are several variations in it. Gray level image enhancement using Modified Random Localization based Differential Evolution (MRLDE) algorithm has been proposed by [\citet{kumar13}]. In this scheme, the variant of DE far better than Genetic Algorithm has been exploited in order to enhance the gray image. However, the convergence is slower compared with Newton-Raphson method since points randomly move about the space and it is required to wait until they move in the right direction. 

The retinex algorithm have been undergone various modifications and improvements, and has been successfully applied to enhance color images of different environmental conditions. Among variety of retinex image enhancement methods presented earlier, NASA's MSR method is popular and have been used widely by other researchers for different applications. It can be seen from the reconstructed images presented by [\citet{Jobson97}] method, there is still considerable room for improvement in the visual quality. The retinex based image enhancement method provides acceptable results if the parameters are adjusted. The selection and tuning of retinex parameters is the most crucial task in a multiscale retinex algorithm. Applying PSO for maximizing the parameters of MSR proposed by [\citet{Hanumantharaju11}] provides better reconstructed images compared with [\citet{Jobson97} method. The core part of the retinex method is to adjust the parameters in such a way that the reconstructed image appears to be visually effective. In contrast to this, applying PSO to maximize parameters of retinex image enhancement method would be an ideal choice to overcome the limitations of retinex based enhancement methods.

\section{Proposed Method}

The most popular retinex based color image enhancement algorithm has reached many improvements and modifications by various researchers. The modified retinex algorithms [\citet{Jharna11},\citet{Liu10}] provide acceptable results and can be further used in image processing system. However, the major concerns with the retinex algorithms are the selection of parameter values. The color image enhancement using MSRCR has many parameters to be considered. For e.g, the parameters such as Gaussian scales $(\sigma_{1}, \sigma_{2}, \sigma_{3})$, number of scales, $\alpha$, $\beta$, G and offset etc., has to be adjusted in order to obtain satisfactory results. The number of parameters and its value selection plays an important role in order to achieve best image enhancement. In this paper, a new approach for the retinex based color image enhancement has been presented. The constant parameters used in the MSRCR algorithm are optimized using the optimization algorithm, PSO. The MSRCR parameters employed for optimization using PSO are Gaussian scales $(\sigma_{1}, \sigma_{2}, \sigma_{3})$, $\alpha$, $\beta$, G and offset etc. The main aim of the proposed method is to find the optimal values for MSRCR parameters. Next, the PSO optimized values for the parameters obtained are used in the MSRCR algorithm to obtain the enhanced image. 

\subsection{Multiscale Retinex with Color Restoration Algorithm}

Conventional image enhancement methods have not fully utilized the human visual characteristics. The retinex based image enhancement is based on the human visual model, in this regard Edwin Land has introduced retinex theory. Land's retinex theory is the basis for the development of the [\citet{Jobson97}] SSR, MSR and MSRCR. The details of the original image are highlighted in the SSR since the image is processed in the logarithmic space. The 2D convolution operation of the original image with Gaussian surround function smoothens the image. Further, the smooth part is subtracted from the original image to obtain the final enhanced image. 

The color image enhancement using SSR can be expressed as

\begin{equation}
R_i(x, y) = \log I_i(x, y) - \log\left[F(x,y)\otimes I_i(x, y) \right]
\end{equation}

where $i\in{R, G, B}, R_i(x, y)$ is the retinex output for the channel i, $I_i(x, y)$ is the image pixel value at location (x, y) for the $i^{th}$ channel, $\otimes$ denotes 2D Gaussian convolution and  $F(x,y)$ is the Gaussian surround function. 

The convolution operation $[F(x,y)\otimes I_i(x,y)]$ represent the illumination component of the $i^{th}$ color spectral band. The retinex output image, R\_i(x,y) is obtained by subtracting the log signal of the illumination component from the original image $I_i(x,y)$  of the $i^{th}$ color spectral band.

The Gaussian surround function F(x,y) is given by Eqn. (2). 

\begin{equation}
F = K\exp\left[\frac{-(x^2+y^2)}{2\sigma^2}\right]
\end{equation}

where x and y are spatial coordinates, $\sigma$ standard deviation of the Gaussian distribution function or Gaussian surround space constant and K is selected such that

\begin{equation}
K = \frac{1}{\sum^{M-1}_{x=0}\sum^{N-1}_{y=0}{F(x, y)}}
\end{equation}

where $M \times N$ represents the size of the image.

The small value of surround space constant results in high dynamic range compression of an image. The large value of Gaussian scale selection produces the enhanced image similar to natural scene. Alternatively, larger the Gaussian scale selection, results in an enhanced image with good color rendition. The middle value of surround space constant is superior for compensating shadow and to accomplish acceptable rendition for the enhanced image. The SSR algorithm improves the luminance of an image without sacrificing the contrast of an image. SSR algorithm is able to provide either dynamic range compression or tonal rendition but not both simultaneously. However, the SSR algorithm introduces halo artifacts in the enhanced images and thus the visual quality of an image degrades. Further, the strength of various surround space constants in the Gaussian function is combined together in MSR algorithm. Therefore, the MSR output is the weighted sum of the several different SSR outputs. 

Mathematically MSR can be represented as Eqn. (4).
\begin{equation}
F_{MSR_{i}} = \sum^{N}_{i=1}W_n \times R_{n_{i}}
\end{equation}

where n is the number of scales assumed as 3 [8] since this matches with the number of color components, $W_n$ is the weighting factor assumed as 1/3 [26], $R_{n_{i}}$ is the SSR output of the $i^{th}$ component of the $n^{th}$ scale and $F_{MSR_{i}}$ is the MSR output of the $i^{th}$ spectral component.
	
The MSRCR is given by Eqn (5):
\begin{equation}
F'_{MSRCR_{i}}(x, y) = C_{i}(x, y)\times F_{MSR_{i}}
\end{equation}

where $i \in {R, G, B}$ and $C_{i}(x,y)$ is expressed as follows:

\begin{equation}
C_{i}(x, y) = \beta \left\{\log \left[\alpha I_{i}(x, y)\right] - \log\left[\sum^{n}_{i=1}I_{i}(x, y)\right]\right\}
\end{equation}

where $\beta$ is a gain and $\alpha$ is a parameter which  indicates the strength of the non-linearity.
The final version of MSRCR is given by Eqn (7):
\begin{equation}
F_{MSRCR_{i}}(x, y) = G\times \left[F'_{MSRCR_{i}}(x, y) + b\right]
\end{equation}
where G and b are the gain and offset values.

\subsection{Objective Criterion}

The quality of an enhanced image is assessed by [\citet{Huang05}, \citet{Huang06}] automatically by employing an efficient objective criterion. The objective function [\citet{Apurba09}, \citet{Kwok06}] used in this work is based on the performance measures such as entropy, sum of edge intensities and number of edge pixels. It can be observed that contrast enhanced image has more number of edge pixels and higher intensity value at the edges. In addition, the entropy value used in the objective criterion reveals the finer details present in the image. 

The objective function is expressed as follows:

\begin{equation}
F(I_e) = \log\left[\log(E(I_s))\right]\times \frac{n\_edgels(I_s)}{MN}\times H(I_{ge})
\end{equation}

where $I_ge$  is the gray-level image of the enhanced color image produced by the MSRCR algorithm. The edges or edgels of Eqn. (8) are determined by using Sobel edge detector. Following the Sobel edge operator, the edge image $I_s$  is obtained for the enhanced gray image. 	$E(I_s)$ represents the sum of MN pixel intensities of Sobel edge image $I_s$ , n\_edgels indicates the number of pixels, whose intensity values is higher than a threshold in the Sobel edge image. Based on the histogram, the entropy value is calculated on the enhanced image as given by Eqn. (9)

\begin{equation}
H(I_{ge}) = -\sum^{255}_{i=0}e_i
\end{equation}

where $e_i= h_{i}log_{2}h_{i}$ , if $hi\neq0$ otherwise $e_i = 0$. The $h_i$ is the probability occurrence of $i^{th}$ intensity value of enhanced gray image $I_{ge}$.

\subsection{PSO Algorithm}

The PSO [\citet{Kwok06}] is a population based search technique modeled on the social behavior of organisms such as bird flocking and fish schooling. The PSO is an algorithm which optimizes the particles within the search space. The individual particles present in the space vary their position with time. In PSO system, particles fly around in a multidimensional search space. During flight, each particle adjusts its position according to its own experience, and the experience of its neighboring particles. Each single solution in the PSO is considered as "particle". All particles have fitness values which are evaluated by the objective function to be optimized and have velocities which direct the flying of the particles. The particles fly through the problem space by following the personal and global best solutions.

The swarm is initialized with a group of random particles and it then searches for optima by updating through iterations. In every iteration, each particle is updated by following two "optima" values. The first one is the best solution of each particle achieved so far. This value is known as "pbest" solution. Another one is the, best solution tracked by any particle among the whole population. This best value is known as "gbest' solution. These two best values are responsible to drive the particles to move to new better position.

After finding the two best values, a particle updates its velocity and position by using the following Eqns. (10) and (11)
\begin{equation}
v_{i}(t+1) = wv_{i}(t)+c_{1}r_{1}\left[p_{i}(t)-x_{i}(t)\right]+c_{2}r_{2}\left[g_{i}(t)-x_{i}(t)\right]
\end{equation}

\begin{equation}
x_{i}(t+1) = x_{i}(t) + v_{i}(t+1)
\end{equation}

where $x_{i}(t)$ and $v_{i}(t)$ indicates the position and velocity of the particle 'i' at time instance 't', w is called inertia weight used to achieve the balance between the global search and local search. c1 and c2 are positive acceleration constants, and $r_1$ and $r_2$ are the random values generated with in the range [0, 1]. $p_{i}(t)$ represents the best solution of the $i^{th}$ particle, and g(t) is the global best solution achieved so far. In Eqn. (10), the first part represents the inertia velocity of the particle, the second part indicates the particle's decision made by its own experience, and the third part signifies the swarm's experience-sociality.

\subsection{Particle Swarm Optimization based Multiscale Retinex with Modified Color Restoration Algorithm}

The proposed PSO based multiscale retinex with color restoration algorithm for image enhancement is presented in this section. The MSR algorithm adapted in the present work is based on modified color restoration technique (MSRMCR) [\citet{Hanumantharaju11}].  The color restoration of MSR method has been modified in such a way that the number of operations are reduced. In order to get the image pixel values in the standard unsigned range of 0 to 255, the multiscale retinex output of Eqn. (4) presented earlier is multiplied by a factor 28.44, followed by a positive offset of 128 as shown in Eqn. (12). 

\begin{equation}
F'_{MSR_{i}} = F_{MSR_{i}}(x, y)\times 28.44 + 128
\end{equation}

This operation translates image pixel values in MSR format from ±4.5 range to [0-255], which is suitable for obtaining visually acceptable picture. Also an additional processing step is carried out using Eqn. (13) in order to solve the gray world violation (which means whitening of pictures). The log(1+x) is used in place of log(x) to ensure a positive result and also to overcome undefined range for log(0). 

\begin{equation}
I'_{i}(x, y) = \log_{2}\left\{1 + C\times\left[\frac{I_{i}(x, y)}{\sum^{n}_{i=1}I_{i}(x, y)}\right]\right\}
\end{equation}

where $i \in {R, G, B}$. In the work of Ref. [\citet{Hanumantharaju11}], a value of 125 is suggested for the constant C. This value empirically settles to 100 for a specific test images.

Further, a scaling factor of $\frac{1}{255}$ is multiplied in order to maintain the value of image in the range of 0 to 255 using Eqn. (14). The choice of scale is application dependent but for most of the applications three scales are required. 

\begin{equation}
F'_{MSRMCR_{i}}(x, y) = \frac{I'_{i}(x, y)\times F'_{MSR_{i}}(x, y)}{255}
\end{equation}

The Multiscale Retinex with Modified Color Restoration (MSRMCR) is computed by using Eqn. (15). The result of the above processing will have both negative and positive RGB values and the histogram will typically have large tails. Thus a final gain-offset is applied to get an enhanced image. The gain of 2.25 and offset value of -30 is used in this method. 

\begin{equation}
F_{MSRMCR_{i}}(x, y) = G\times\left[F'_{MSRMCR_{i}}(x, y) + b\right]
\end{equation}

where $F_{MSRMCR_{i}}(x,y)$  is the output of the multiscale retinex with modified color restoration.

The multiscale retinex parameters such as Gaussian scales ($(\sigma_{1}, \sigma_{2}, \sigma_{3})$), empirical constant (C), gain (G) and offset (b) used in the MSRMCR image enhancement algorithm are optimized using PSO. \\


\begin{figure*}
\caption{Flow Chart of the Proposed PSO based MSRMCR Color Image Enhancement}
\begin{center}$
\begin{array}{c}
\includegraphics[height=9.3 cm, width=12 cm]{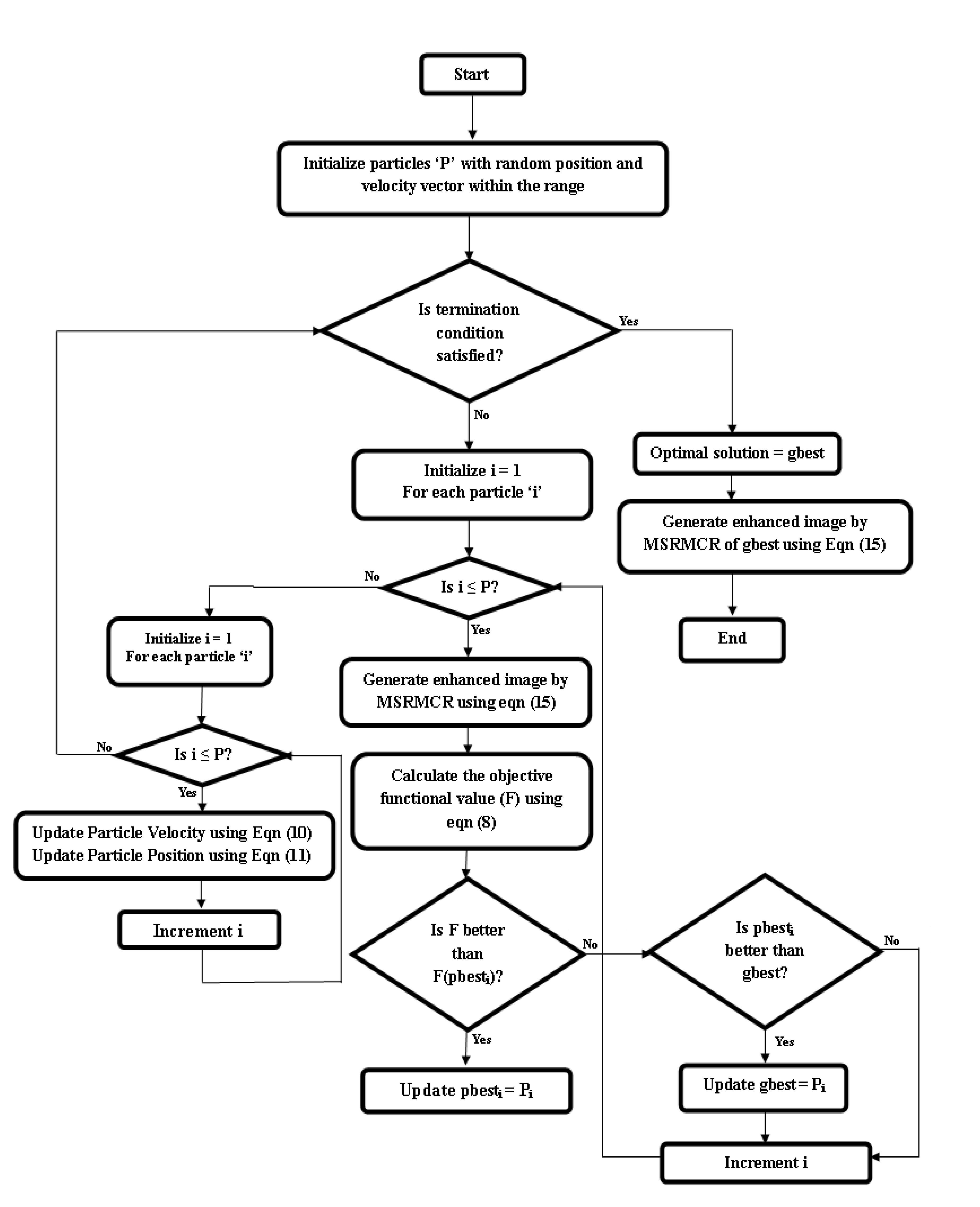} 
\end{array}$
\end{center}
\end{figure*}

The flowchart of the proposed PSO based MSRMCR algorithm is presented in Fig. 1. The parameters chosen in the proposed scheme are as follows: The number of particles 'P' and dimension 'd' of each particle is set to 30 and 6 respectively. Next, we choose the number of iterations as 30 in order to terminate the algorithm. It is clear from the experimental results that the number of iterations chosen in the present work produces satisfactory results. The range preferred in this scheme for Gaussian scales are as follows: $\sigma_{1} \in [1, 80]$, $\sigma_{21} \in [81, 150]$, $\sigma_{3} \in [151, 256]$.  The range of parameters chosen for NASA's MSRCR using PSO are $\alpha \in [100, 125]$, $G \in [150, 200]$ and $\beta \in [0, 50]$ respectively. The range of values selected for proposed MSRMCR parameters are as follows: C, G and b are selected with [100, 125], [0, 5] and [-50, 0] respectively. The random numbers r1 and r2 used in Eqn. (10) are selected between 0 and 1. The positive acceleration constants ac1 and ac2 are also random numbers selected in the range of [0, 2].

The inertia weight W which decreases linearly with the increase of generation is given by Eqn. (16).

\begin{equation}
W = W_{max} - Current_{iteration}\times \frac{W_{max}-W_{min}}{max_{iteration}}
\end{equation}

where $W_{max}$, is the largest weighting coefficient which is set to 2, the $W_{min}$ is the minimum weighted coefficient, is set to 0, $Current_{iteration}$ is the current iteration and $max_{iteration}$ is the total number of iterations set for the algorithm. 

\section{Experimental Results and Comparative Study}

This section presents the experimental results of the proposed PSO based MSRMCR image enhancement presented in the earlier section. In this paper, the most popular image enhancement techniques namely, HE [\citet{Lu10}, \citet{Sengee10}], NASA's MSRCR, IMSRCR [\citet{Chih09}] and photoflair based software are chosen in order to validate the proposed scheme. The photoflair is photo enhancement software by Truview imaging company who patented retinex image enhancement algorithm from NASA. Photoflair restores rich colors in an image by correcting underexposed dark areas without affecting correctly exposed areas.

The work presented here is validated by considering number of test images with resolution of $512\times512$ pixels. The input images are resized to $256\times256$ pixels in order to reduce execution time and are considered in Tagged Image File Format (.tif) due to its advantage of platform independent and storage requirement. The image enhancement algorithms mentioned earlier are invoked to compare with the proposed method. The algorithm have been coded and tested using Matlab version 8.0. Elaborate experiments were conducted on over a dozen varieties of images and consistently good results have been obtained for the proposed method. As examples, a few poor quality images have been enhanced using the proposed method and is presented. Table 1 provides the [\citet{Jobson97}] MSRCR parameters tuned for first set of test images using PSO algorithm. Table 2 shows the proposed MSRMCR parameters obtained for the first set of test images based on PSO algorithm. The tuned parameters are embedded into the MSRCR and MSRMCR image enhancement algorithms, respectively. 
	
\begin{table}
\renewcommand{\arraystretch}{1.3}
\caption{MSRCR [\citet{Jobson97}] Parameters Tuned using PSO Algorithm}
\label{table_example}
\centering
\begin{tabular}{|l|l|l|l|l|l|l|l|l|}
\hline
\bf Image & \bf Resolution & \bf $\sigma_{1}$ & $\sigma_{2}$ & $\sigma_{3}$ & G & $\alpha$ & $\beta$& b \\
\hline
Office    &  $256\times256$ & 52.55  & 83.65 & 198.84 & 168 & 121 & 38 & -12  \\
\hline
Tree      &  $256\times256$ & 18.63  &139.90  &162.72  & 153 &137  &45 & -17 \\
\hline
Girl      &  $256\times256$ & 29.63  & 97.22 & 187.87 &167 & 129 & 49& -20 \\
\hline
Bird      &  $256\times256$ & 20.21   &86.62  & 221.20 &186  &131  &42 & -18  \\
\hline
\end{tabular}
\end{table}

\begin{table}
\renewcommand{\arraystretch}{1.3}
\caption{Proposed MSRMCR Parameters Tuned using PSO Algorithm}
\label{table_example}
\centering
\begin{tabular}{|l|l|l|l|l|l|l|l|}
\hline
\bf Image & \bf Resolution & \bf $\sigma_{1}$ & $\sigma_{2}$ & $\sigma_{3}$ & C & G & b \\
\hline
Office    &  $256\times256$ & 63.83	&128.60	&201.08	&102&	2.21&	-13  \\
\hline
Tree      &  $256\times256$ & 21.93	&147.37	&158.34	&122	&3.15&	-12 \\
\hline
Girl      &  $256\times256$ & 37.74	&94.56	&175.85	&108	&1.98&	-21 \\
\hline
Bird      &  $256\times256$ & 24.87	&121.62	&205.64&	120	&2.25&	-14  \\
\hline
\end{tabular}
\end{table}

The original 'office' image with office light off, intense background and dark foreground is shown in Fig. 2(a). The reconstructed images using HE [\citet{Cheng04}], NASA's MSRCR of Ref. [\citet{Jobson97}], Improved MSR of Ref. [\citet{Chih09}], Photoflair and proposed method were shown in Figs. 2(b), (c), (d) (e) and (f), respectively. As is evident from the results presented, the proposed method yields better enhanced color images, since the use of PSO optimizes the parameters. The swarm intelligence is an innovative paradigm for solving optimization problems in image enhancement. The PSO incorporated into retinex method and uses the maximized parameters adaptively. The PSO is a population-based optimization method, which can be implemented and applied easily to solve various optimization problems of image enhancement. The main strength of the PSO algorithm is its fast convergence, compared with GA, SA and DE based optimization algorithms. In this paper, the PSO is used to investigate the optimal parameters for MSR algorithm in order to obtain the best image enhancement. Initially, the PSO algorithm randomly selects the number of particles and parameters within their range. During each iteration of the algorithm, each particle is evaluated by the objective function, determining the fitness of the solution. The PSO approach used in this work for color image enhancement not only selects optimal parameters but also evaluates the enhanced image based on entropy, edge information of an image and number of edge pixels.

\begin{table}
\renewcommand{\arraystretch}{1.3}
\caption{Comparison of Proposed MSRCR Image Enhancement Method with NASA's MSRCR based on Computation Complexity for 200 Generations}
\label{table_example}
\centering
\begin{tabular}{|l|c|c|c|}
\hline
\bf Image & \bf Resolution & Proposed MSRMCR (Sec) & NASA's MSRCR (Sec) \\
\hline
Office    &  $256\times256$ & 225  & 283.5\\
\hline
Tree      &  $256\times256$ & 232.3  & 297.1\\
\hline
Girl      &  $256\times256$ & 221.9  & 271.6 \\
\hline
Bird      &  $256\times256$ & 229.2  &  288\\
\hline
\end{tabular}
\end{table}

\begin{figure*}[ht]
\caption{Comparisons of Reconstructed 'office' Images Using Image Enhancement Algorithms (a) Original Image (b) Image Enhanced using HE of Ref. [\citet{Cheng04}] (c) Reconstructed Image using the method proposed in Ref. [\citet{Jobson97}] (d) Improved MSR of Ref. [\citet{Chih09}] (e) Photoflair Software based Enhancement (f) Proposed PSO based MSRMCR method. \textbf{Note:} The 'office' image with office light off is from http://ivrg.epfl.ch/index.html.}
\begin{center}$
\begin{array}{c}
\includegraphics[height=7 cm, width=10 cm]{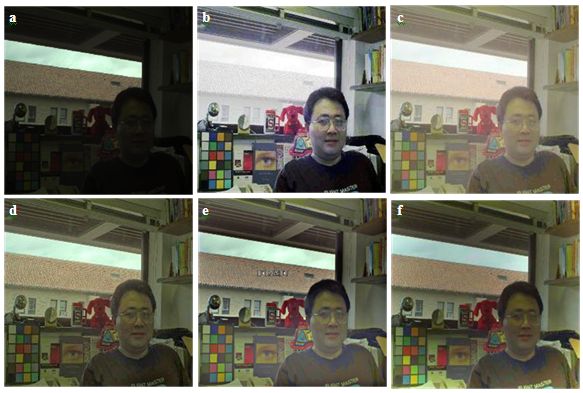} 
\end{array}$
\end{center}
\end{figure*} 

The proposed algorithm is tested with other images such as 'tree', 'girl', \&\ 'bird' and are presented in Figs. 3, 4 and 5, respectively. As is evident from the Matlab simulation results, the proposed PSO based MSRMCR enhancement method provides better reconstruction compared to MSRCR method of Ref. [\citet{Jobson97}]. Referring to 'Office' image, it can be observed that the computational time for the proposed PSO based MSRMCR was found to 225 sec. The time taken for NASA's retinex of Ref. [\citet{Jobson97}] was found 283.5 sec. This is due to the fact that the PSO based MSRMCR algorithm has less number of computations. This drastically reduces the computation time of the algorithm, whereas applying PSO for MSRCR increases the processing time considerably. The NASA's MSRCR algorithm is computationally intensive and demands high execution time. Table 3 provides comparison of the proposed MSRMCR image enhancement scheme with the NASA's MSRCR method based on the computation time.          

\begin{figure}
\caption{Comparisons of Reconstructed 'tree' Images Using Image Enhancement Algorithms (a) Original Image (b) Image Enhanced using HE of Ref. [\citet{Cheng04}] (c) Reconstructed Image using the method proposed in Ref. [\citet{Jobson97}] (d) Improved Multiscale Retinex of Ref. [\citet{Chih09}] (e) Photoflair Software based Enhancement (f) Proposed PSO based MSRMCR method. \textbf{Note:} The 'office' image with office light off is from http://ivrg.epfl.ch/index.html. \textbf{Note:}The original 'tree' image is from http://ivrg.epfl.ch/index.html.
}
\begin{center}$
\begin{array}{c}
\includegraphics[height=7 cm, width=10 cm]{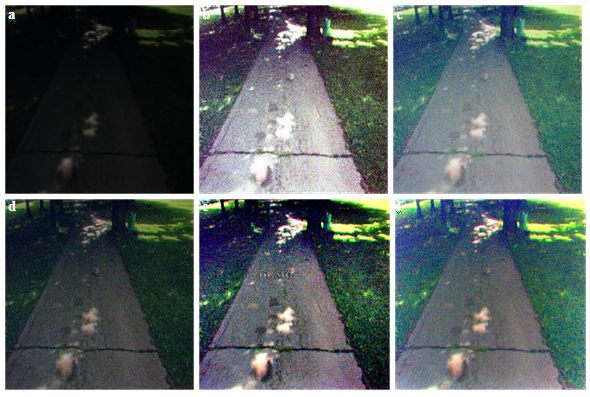} 
\end{array}$
\end{center}
\end{figure} 

\begin{figure}
\caption{Comparisons of Reconstructed 'girl' Images Using Image Enhancement Algorithms (a) Original Image (b) Image Enhanced using HE of Ref. [\citet{Cheng04}] (c) Reconstructed Image using the method proposed in Ref. [\citet{Jobson97}] (d) Improved MSR of Ref. [\citet{Chih09}] (e) Photoflair Software based Enhancement (f) Proposed PSO based MSRMCR method. \textbf{Note:}The original 'girl' image is from http://dragon.larc.nasa.gov/retinex/pao/news/}
\begin{center}$
\begin{array}{c}
\includegraphics[height=7 cm, width=10 cm]{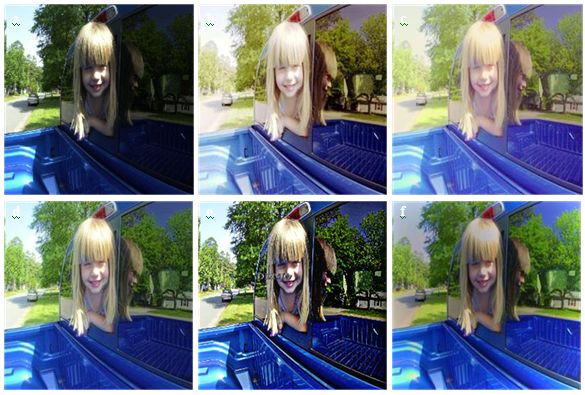} 
\end{array}$
\end{center}
\end{figure}

\begin{figure}
\caption{Comparisons of Reconstructed 'bird' Images Using Image Enhancement Algorithms (a) Original Image (b) Image Enhanced using HE of Ref. [\citet{Cheng04}] (c) Reconstructed Image using the method proposed in Ref. [\citet{Jobson97}] (d) Improved MSR of Ref. [\citet{Chih09}] (e) Photoflair Software based Enhancement (f) Proposed PSO based MSRMCR method. \textbf{Note:}The original 'bird' image is from http://www.cs.huji.ac.il/~danix/hdr/enhancement.html}
\begin{center}$
\begin{array}{c}
\includegraphics[height=7 cm, width=10 cm]{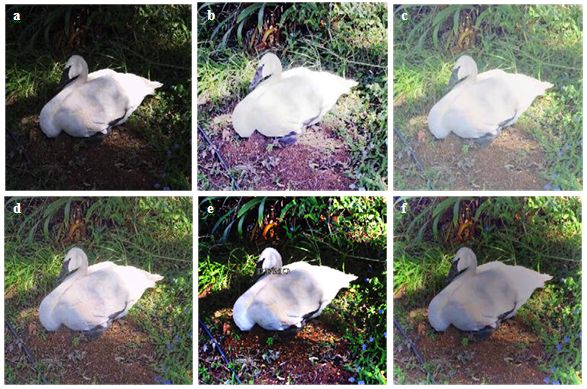} 
\end{array}$
\end{center}
\end{figure} 
 
In order to validate the performance of the proposed algorithm with more details, additional experimental results of 'Aerial', 'Palace', 'Big-ben', 'Memorial Church' and 'House' images are presented in Figs. 6, 7, 8, 9 and 10, respectively. The Table 4 provides the parameters adjusted using PSO for MSRCR method of Ref. [\citet{Jobson97}]. Table 5 shows the proposed MSRMCR parameters obtained for second set of test images based on PSO algorithm. The tuned parameters are embedded into the MSRCR and MSRMCR image enhancement algorithms, respectively.  It may be observed from the results presented that the proposed MSRMCR method offers good enhancement results far better than the other researcher methods. Table 6 provides computational complexity of the reconstructed images using proposed MSRMCR and NASA's MSRCR image enhancement techniques for additional test images.  

\begin{table}
\renewcommand{\arraystretch}{1.3}
\caption{MSRCR [\citet{Jobson97}] Parameters Tuned using PSO Algorithm for Additional Test Images}
\label{table_example}
\centering
\begin{tabular}{|l|l|l|l|l|l|l|l|l|}
\hline
\bf Image & \bf Resolution & \bf $\sigma_{1}$ & $\sigma_{2}$ & $\sigma_{3}$ & G & $\alpha$ & $\beta$& b \\
\hline
Aerial     &  $256\times256$ & 13.18&	99.01	&177.22&	189&	113&	39&	-25  \\
\hline
Palace     &  $256\times256$ & 20.19&	127.82&	192.81&	133&	111&	33&	-27 \\
\hline
Big-ben    &  $256\times256$ & 67.51&	119.87&	201.01&	161&	117&	44&	-19 \\
\hline
Memorial   &  $256\times256$ & 57.99&	134.15&	197.19&	198&	119&	36&	-15  \\
\hline
House      &  $256\times256$ & 51.20&	116.90&	216.81&	173&	122&	31&	-12  \\
\hline
\end{tabular}
\end{table}

\begin{table}
\renewcommand{\arraystretch}{1.3}
\caption{Proposed MSRMCR Parameters Tuned using PSO Algorithm for Additional Test Images}
\label{table_example}
\centering
\begin{tabular}{|l|l|l|l|l|l|l|l|}
\hline
\bf Image & \bf Resolution & \bf $\sigma_{1}$ & $\sigma_{2}$ & $\sigma_{3}$ & C & G & b \\
\hline
Aerial   &  $256\times256$ & 13.18&	99.01&	177.22&	119&	2.12&	-25  \\
\hline
Palace      &  $256\times256$ & 20.19&	127.82&	192.81&	113&	1.92&	-27 \\
\hline
Big-ben      &  $256\times256$ & 67.51&	119.87&	201.01&	101&	2.25&	-19 \\
\hline
Memorial      &  $256\times256$ & 57.99&	134.15&	197.19&	111&	1.29&	-15  \\
\hline
House      &  $256\times256$ & 51.20&	116.90&	216.81&	113&	1.11&	-12  \\
\hline
\end{tabular}
\end{table}

\begin{table}
\renewcommand{\arraystretch}{1.3}
\caption{Comparison of Proposed MSRCR Image Enhancement Method with NASA's MSRCR based on Computation Complexity for 200 Generations}
\label{table_example}
\centering
\begin{tabular}{|l|c|c|c|}
\hline
\bf Image & \bf Resolution & Proposed MSRMCR (Sec) & NASA's MSRCR (Sec) \\
\hline
Aerial    &  $256\times256$ & 231.7  & 289.1\\
\hline
Palace      &  $256\times256$ & 229.8  & 299.5\\
\hline
Big-Ben      &  $256\times256$ & 232.7  & 287.3 \\
\hline
Memorial Church      &  $256\times256$ & 233.2  &  294.2\\
\hline
House      &  $256\times256$ & 232.9  &  288.6\\
\hline
\end{tabular}
\end{table}

\begin{figure}
\caption{Image Enhancement Methods comparison
(a) Original 'Aerial' Image of resolution $256\times256$ pixels (b) Image Enhanced based on Histogram Equalization of Ref. [\citet{Cheng04}] (c) Image Enhanced using NASA's MSRCR technology of Ref. [\citet{Jobson97}] (d) Improved MSRCR based Color Image Enhancement of Ref. [\citet{Chih09}] (e) Image Enhanced using Proposed PSO based MSRMCR Note:  The original 'aerial' image is from http://dragon.larc.nasa.gov/retinex/pao/news/}
\begin{center}$
\begin{array}{c}
\includegraphics[height=7 cm, width=10 cm]{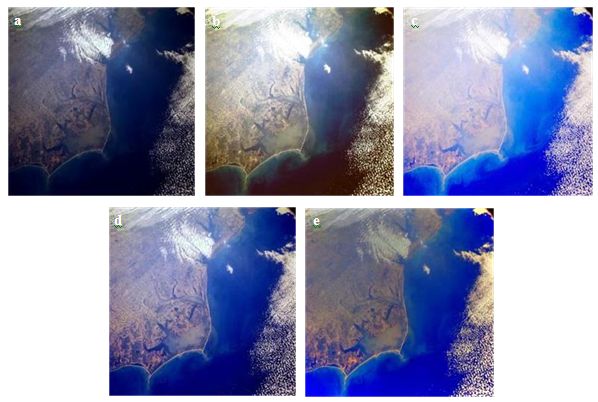} 
\end{array}$
\end{center}
\end{figure} 

\begin{figure}
\caption{Image Enhancement Methods comparison (a) Original 'Palace' Image of resolution $256\times256$ pixels (b) Image Enhanced based on Histogram Equalization of Ref. [\citet{Cheng04}] (c) Image Enhanced using NASA's MSRCR technology of Ref. [\citet{Jobson97}] (d) Improved MSRCR based Color Image Enhancement of Ref. [\citet{Chih09}] (e) Image Enhanced using Proposed PSO based MSRMCR
Note:  The original 'palace' image is from http://visl.technion.ac.il/1999/99-07/www/}
\begin{center}$
\begin{array}{c}
\includegraphics[height=7 cm, width=10 cm]{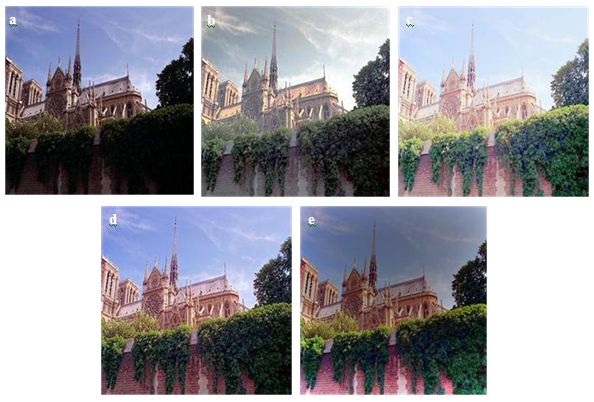} 
\end{array}$
\end{center}
\end{figure} 

\begin{figure}
\caption{Image Enhancement Methods comparison (a) Original 'Big-ben' Image of resolution $256\times256$ pixels (b) Image Enhanced based on Histogram Equalization of Ref. [\citet{Cheng04}] (c) Image Enhanced using NASA's MSRCR technology of Ref. [\citet{Jobson97}] (d) Improved MSRCR based Color Image Enhancement of Ref. [\citet{Chih09}] (e) Image Enhanced using Proposed PSO based modified MSRCR Note: The original 'Big-ben' image is from http://visl.technion.ac.il/1999/99-07/www/}
\begin{center}$
\begin{array}{c}
\includegraphics[height=7 cm, width=10 cm]{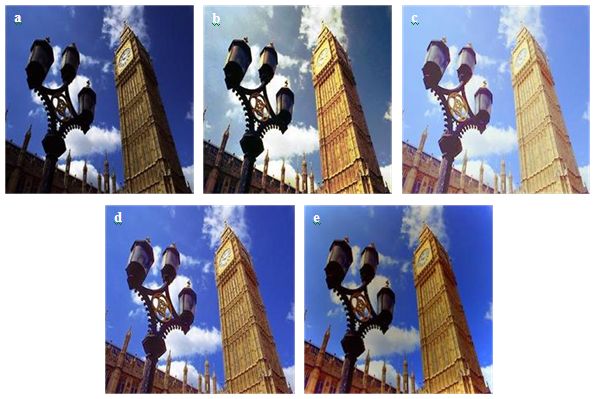} 
\end{array}$
\end{center}
\end{figure}

\begin{figure*}[ht]
\caption{Image Enhancement Methods comparison
(a) Original 'Memorial Church' Image of resolution $256\times256$ pixels (b) Image Enhanced based on Histogram Equalization of [\citet{Cheng04}] (c) Image Enhanced using NASA's MSRCR technology of Ref. [\citet{Jobson97}] (d) Improved MSRCR based Color Image Enhancement of Ref. [\citet{Chih09}] (e) Image Enhanced using Proposed PSO based MSRMCR
Note:  The original 'memorial church' image is from http://ivrg.epfl.ch/index.html.
}
\begin{center}$
\begin{array}{c}
\includegraphics[height=7 cm, width=10 cm]{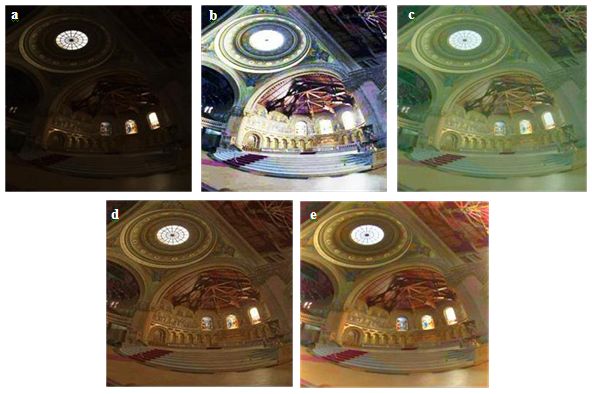} 
\end{array}$
\end{center}
\end{figure*}	

\begin{figure*}[ht]
\caption{Image Enhancement Methods comparison
(a) Original 'house' Image of resolution $256\times256$ pixels (b) Image Enhanced based on Histogram Equalization of Ref. [\citet{Cheng04}] (c) Image Enhanced using NASA's MSRCR technology of Ref. [\citet{Jobson97}] (d) Improved MSRCR based Color Image Enhancement of Ref. [\citet{Chih09}] (e) Image Enhanced using Proposed PSO based MSRMCR
}
\begin{center}$
\begin{array}{c}
\includegraphics[height=7 cm, width=10 cm]{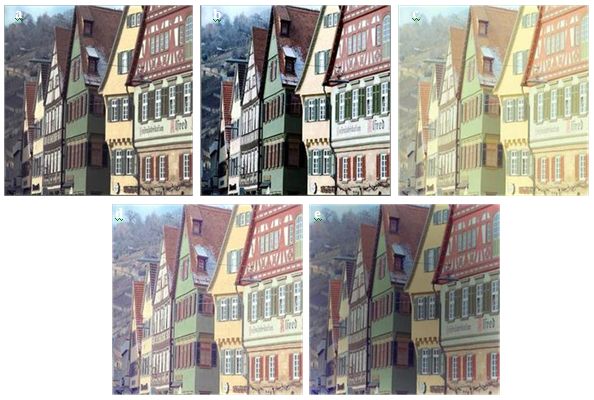} 
\end{array}$
\end{center}
\end{figure*}	

The performance of the proposed enhancement algorithm is compared with other enhancement schemes described earlier based on the new metric called WE [\citet{raju11}]. The WE is computed for original image and enhanced image subsequent to two dimensional Discrete Wavelet Transform (2D-DWT). The Approximate WE (AWE) and Detailed WE (DWE) computed are used for evaluating the quality of the enhanced image. In this paper, AWE and DWE adapted for image quality assessment provides information about global appearance and local details of an image. The Fig. 11 and Fig. 12 shows the comparison of the proposed PSO-MSRMCR with other image enhancement algorithms using a WE metric. The developed enhancement algorithm not only improves luminance and color fidelity but also adapts Human Vision System (HVS) characteristics. In our approach, the AWE and DWE values of the best enhanced image are to be equal to the AWE and DWE values of the original image. The proposed WE metric exploited for image quality assessment make sure that color and details of the image are better for human vision perception. 

\begin{figure}
\caption{Comparison of the proposed PSO-MSRMCR result with other image enhancement algorithms using a WE Metric: (a) Approximate WE Metric Comparison for Various Test Images. (b) Detailed WE Metric Comparison for Various Test Images. \textbf{Note:} WE metric for test images 'Office', 'Tree', 'Girl' and 'Bird', respectively.}
\begin{center}$
\begin{array}{cc}
\includegraphics[height=7 cm, width=10 cm]{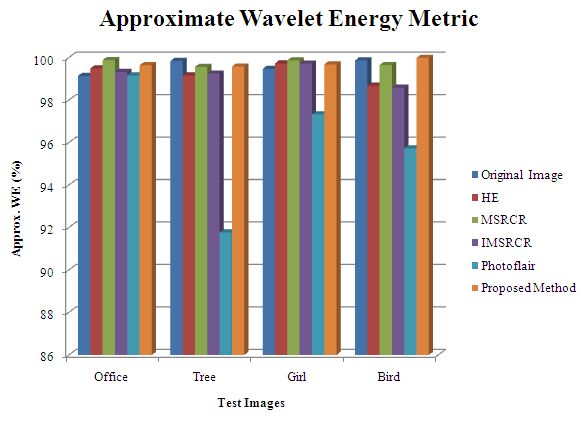} \\
\includegraphics[height=7 cm, width=10 cm]{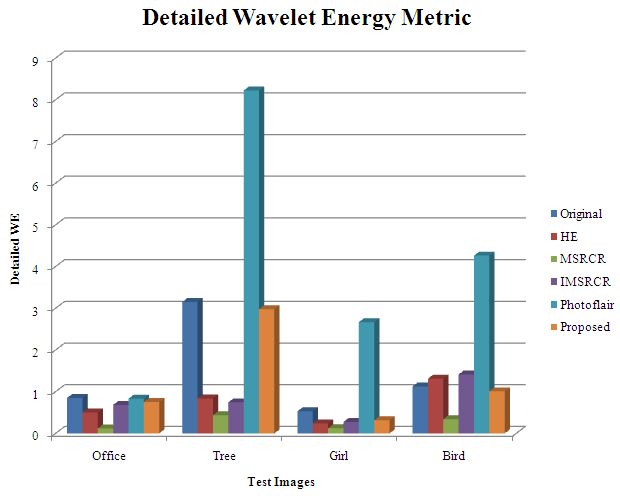}
\end{array}$
\end{center}
\end{figure}	

\begin{figure}
\caption{Comparison of the proposed PSO-MSRMCR result with other image enhancement algorithms using a WE Metric: (a) Approximate WE Metric Comparison for Additional Test Images. (b) c.	Detailed WE Metric Comparison for Additional Test Images. \textbf{Note:} WE metric for test images 'Aerial', 'Palace', 'Big-ben', 'Memorial' and 'House' respectively.}
\begin{center}$
\begin{array}{cc}
\includegraphics[height=7 cm, width=10 cm]{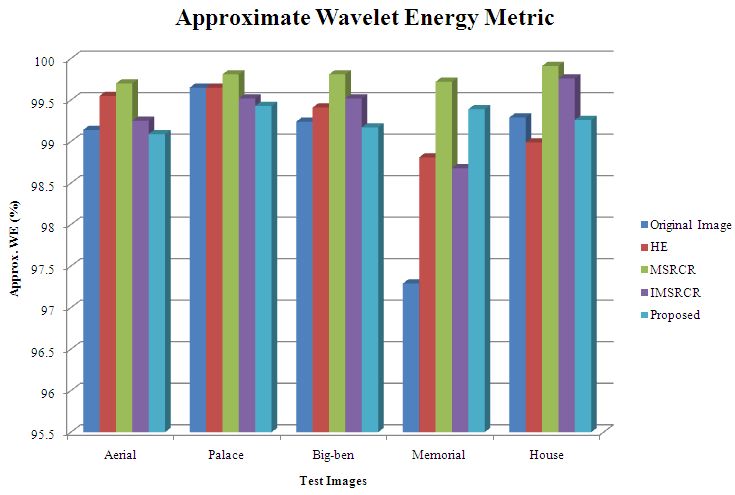} \\
\includegraphics[height=7 cm, width=10 cm]{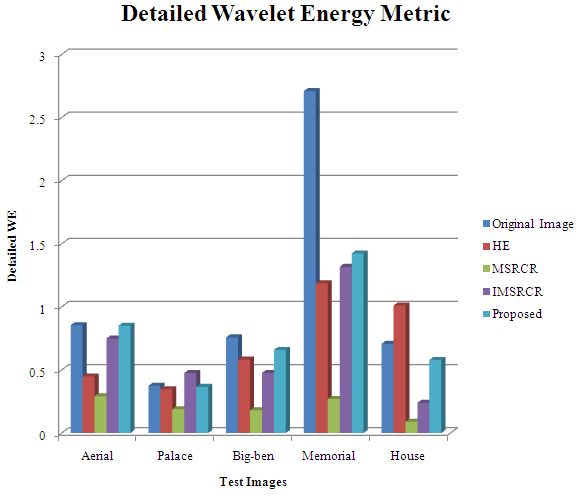}
\end{array}$
\end{center}
\end{figure}	

It is clear from the experimental results that the proposed algorithm outperforms the other image enhancement techniques in terms of quality of the reconstructed picture. The images enhanced by our method are clearer, more vivid, and more brilliant than that achieved by other researchers. 

\section{Conclusion}

A new approach for color image enhancement using PSO based MSRMCR has been presented in this paper. The parameters selection in an MSRMCR based color image enhancement algorithm was a critical task. Due to the hard selection of parameters, PSO algorithm has been used in order to obtain the optimal parameters. The PSO algorithm exploited in the proposed scheme utilizes an efficient objective criterion based on entropy and edge information of an image. The MSR parameters such as Gaussian surround space constants, number of scales, gain and offset etc. are optimized using PSO technique to obtain reconstructed images far better than other researcher methods. The visual quality of the reconstructed images are evaluated using a new WE metric. The Matlab code developed for the proposed work was tested with various images of different environmental conditions and found to produce high quality enhanced images. Currently, work is under progress for amending the algorithm for facial images, medical images etc.



%
%
%
%



%

\end{document}